\documentclass[letterpaper, 10 pt, conference]{ieeeconf}  

\IEEEoverridecommandlockouts                              

\usepackage{graphics} 
\usepackage{subfig} 
\usepackage{threeparttable} 
\usepackage{multirow,tabularx} 
\usepackage{makecell} 
\usepackage{amsmath} 

\usepackage{epsfig} 
\usepackage{amsmath} 
\usepackage{amssymb}  

\usepackage[normalem]{ulem}
\usepackage{csquotes}
\usepackage{hyperref}
\hypersetup{
    colorlinks=true,
    linkcolor=blue,
    urlcolor=blue,
    citecolor=cyan,
}
\usepackage{color}

\usepackage{xcolor}

\title{\LARGE \bf
Satellite Image Based Cross-view Localization for Autonomous Vehicle
}

\author{ 
Shan Wang$^{1,2}$, Yanhao Zhang$^{1}$, Ankit Vora$^{3}$, Akhil Perincherry$^{3}$, and Hongdong Li$^{1}$
\thanks{$^{1}$ Shan Wang, Yanhao Zhang, and Hongdong Li are with Australian National University.
{\tt\small Shan.Wang@anu.edu.au.}}%
\thanks{$^{2}$ Shan Wang is also with Data61, CSIRO, Australia.}%
\thanks{$^{3}$ Ankit Vora and Akhil Perincherry are with Ford Motor Company, Dearborn, USA.}%
}

\begin{document}

\maketitle
\thispagestyle{empty}
\pagestyle{empty}

\begin{abstract}
Existing spatial localization techniques for autonomous vehicles mostly use a pre-built 3D-HD map, often constructed using a survey-grade 3D mapping vehicle, which is not only expensive but also laborious.  This paper shows that by using an off-the-shelf high-definition satellite image as a ready-to-use map, we are able to achieve cross-view vehicle localization up to a satisfactory accuracy, providing a cheaper and more practical way for localization.  
While the utilization of satellite imagery for cross-view localization is an established concept, the conventional methodology focuses primarily on image retrieval. This paper introduces a novel approach to cross-view localization that departs from the conventional image retrieval method.
Specifically, our method develops (1) a Geometric-align Feature Extractor (GaFE) that leverages measured 3D points to bridge the geometric gap between ground and overhead views, (2) a Pose Aware Branch (PAB) adopting a triplet loss to encourage pose-aware feature extraction, and (3) a Recursive Pose Refine Branch (RPRB) using the Levenberg-Marquardt (LM) algorithm to align the initial pose towards the true vehicle pose iteratively. Our method is validated on KITTI and Ford Multi-AV Seasonal datasets as ground view and Google Maps as the satellite view. The results demonstrate the superiority of our method in cross-view localization with median spatial and angular errors within $1$ meter and $1^\circ$, respectively. 
\end{abstract}

\begin{keywords}
Cross-View localization, Pose Estimation, Deep Learning
\end{keywords}

\section{Introduction}
Accurate vehicle localization plays an enabling role in autonomous driving. Although consumer-grade GPS devices have been widely used for vehicle localization, their performances degrade rapidly in GPS-compromised areas \cite{xiong2021g}. For instance, it is difficult to obtain reliable localization in urban areas with high-rise buildings. 
Other sensor modalities such as camera \cite{mcmanus2014shady,pascoe2015direct,pascoe2015robust}, and LiDAR \cite{maddern2015leveraging,le2019in2lama,8461000,voraAerial} are explored for achieving robust vehicle localization. Yet, the existing vehicle localization techniques critically rely on a pre-built 3D high-definition map. Both the acquisition and maintenance of such a 3D HD map are laborious and expensive, especially for rural areas where a mapping vehicle only visits rather infrequently.
Cross-view Localization, by using off-the-shelf commercially-available satellite images as a map in spatial accordance with ground-view images captured by vehicle cameras, provides a cost-effective and promising solution. Recently, several works have been published in this front \cite{shi2020optimal,shi2019spatial,hu2018cvm,liu2019lending,toker2021coming}. These cross-view localization methods, however, do not take full advantage of the satellite information. Instead, they only handle it via the conventional image-retrieval idea, hence they only achieve coarse localization. 

\begin{figure}
    \centering
    \begin{minipage}[b]{0.45\textwidth}
    \subfloat[\it Query]{
    \includegraphics[width=0.595\linewidth]{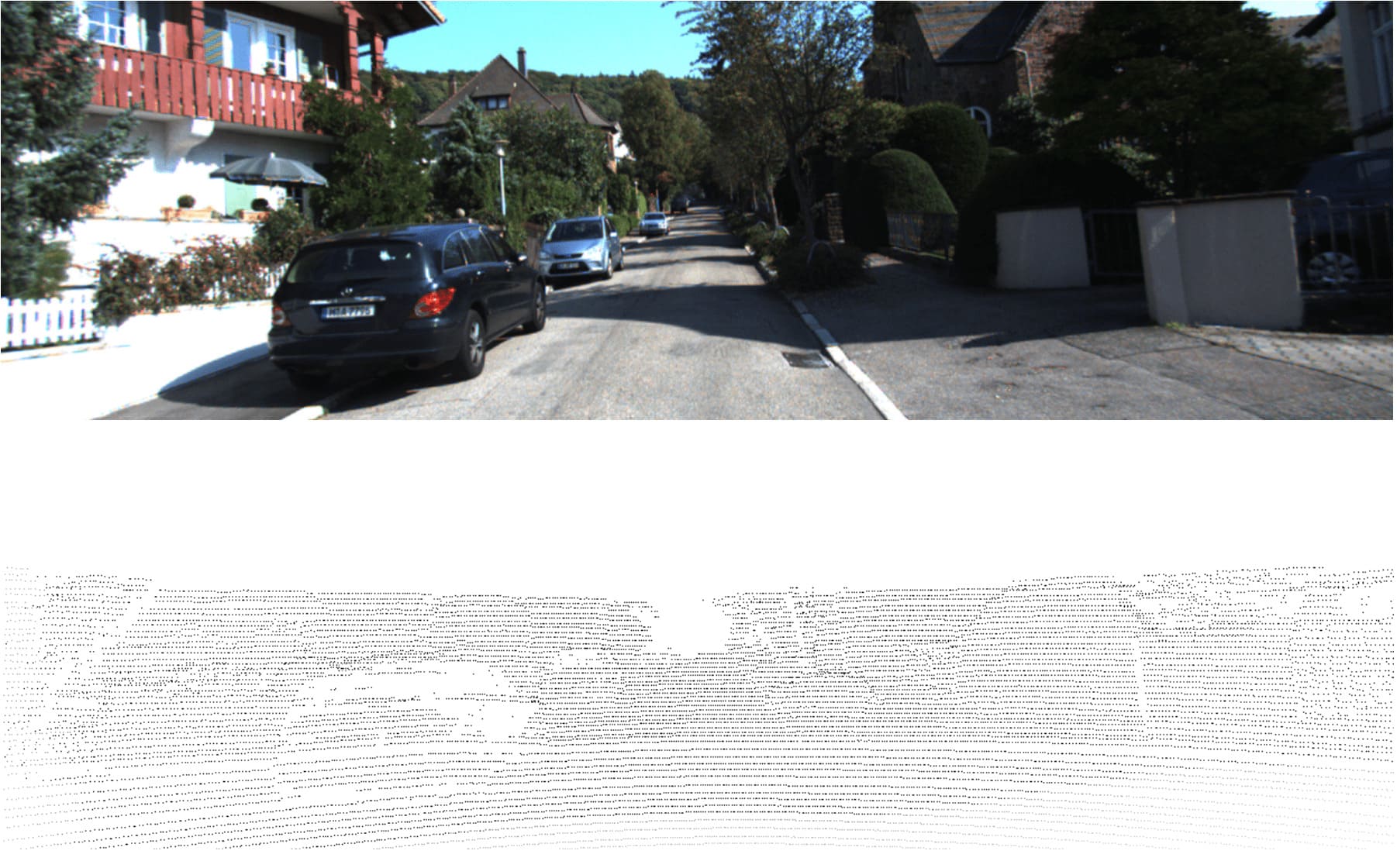}
    }
    \subfloat[\it Reference]{
    \includegraphics[width=0.37\linewidth]{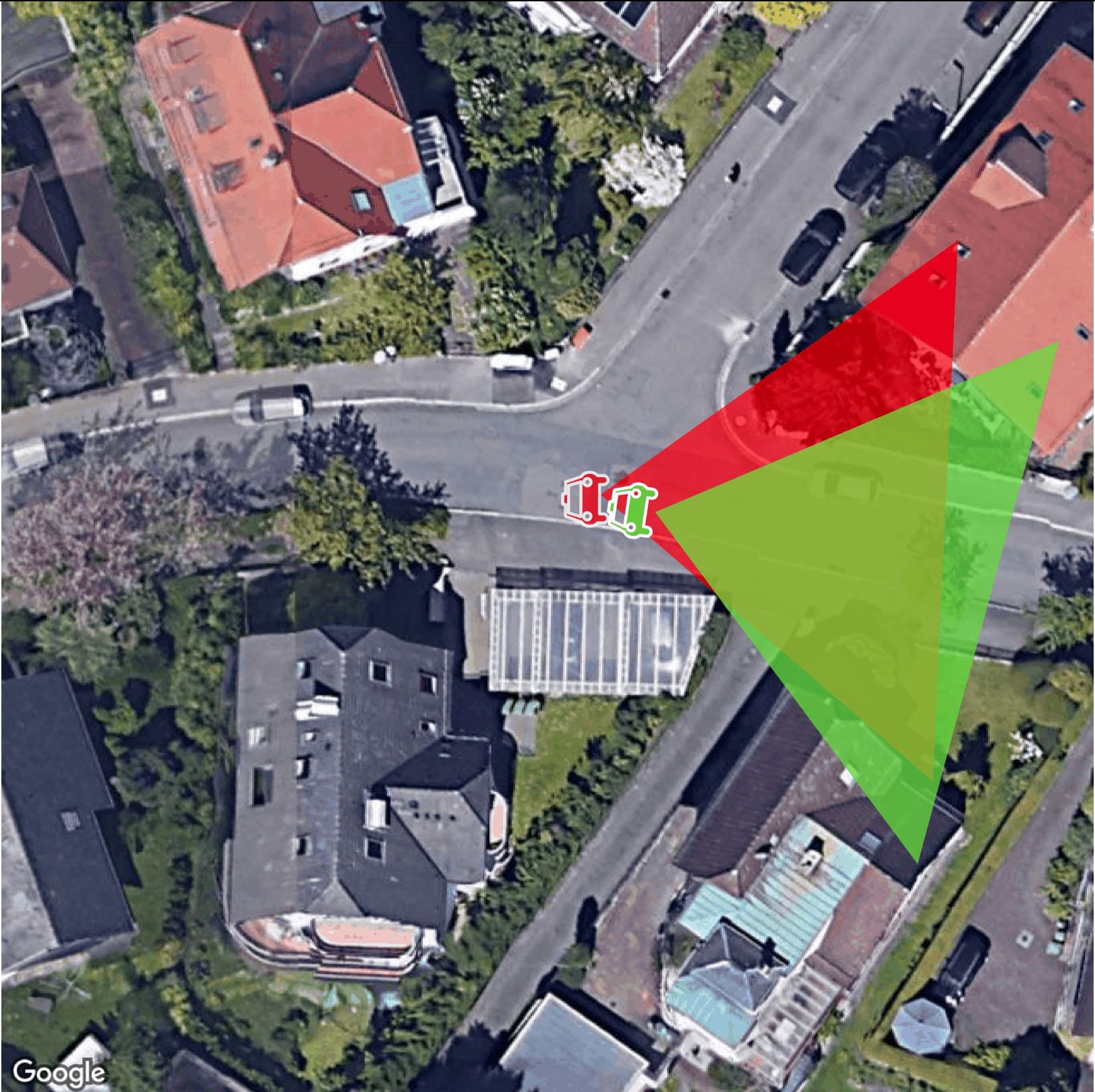}
    }
    \end{minipage}
    \caption{(a) Query information, including a ground-view image and corresponding 3D LiDAR points, and (b) reference satellite image. We aim to estimate the accurate 3-DoF pose of the ground-view camera. The initial and accurate 3-DoF poses are shown in \textcolor{red}{(red)} and \textcolor{green}{(green)}, respectively.}
    \label{fig:Task}
    \vspace{-0.2in}
\end{figure}

Different from the image-retrieval-based cross-view methods, we propose a new fine-grained cross-view localization in this paper. Given a spatially-consistent satellite image, our method aims to estimate the accurate 3-DoF pose of the vehicle using a ground-view image (vehicle camera) and the 3D LiDAR points. 
Fig.~\ref{fig:Task} illustrates the setting of our method. 
To mitigate the domain gap between ground-view images and satellite images, we establish correspondences between the two views by projecting 3D LiDAR points onto their respective images.
Given an initial coarse pose on the satellite map, we iteratively optimize the pose through feature matching.

Our Satellite Image Based Cross-view Localization (SIBCL) deep neural network consists of a Geometric-align Feature Extractor (GaFE) and two branches of objective functions, Pose-Aware Branch (PAB) and Recursive Pose Refine Branch (RPRB). More specifically, GaFE embeds ground-view and satellite images into the feature space with a shared weights encoder. It further establishes spatial-feature correspondences by projecting the 3D points onto the respective images. 
PAB employs a triplet loss \cite{qian2019softtriple} to differentiate the variation across point features (residual)
between two views conditioned on the correct (ground truth) and incorrect (initial) pose. 
RPRB is tasked to iteratively optimize the initial pose towards the ground truth pose with the Levenberg–Marquardt (LM) algorithm. Moreover, a re-projection error is deployed on the optimized pose. It is noted that both objective branches supervise feature extraction but have a different focus. PAB encourages the correct pose estimation as well as penalizes for the incorrect. RPRB encourages the most correct (predict) pose close to the ground truth.

In order to train and evaluate our method, we construct two cross-view localization datasets, KITTI-CVL and FordAV-CVL.
These are composed of ground-view images from KITTI \cite{geiger2013vision}, Ford Multi-AV Seasonal \cite{Agarwal_2020} datasets respectively, and their spatial-consistent satellite counterparts from Google Map \cite{google} according to image-wise GPS information. 
Extensive experiments on the proposed KITTI-CVL and FordAV-CVL datasets demonstrate that our SIBCL can accurately estimate vehicle positions with median errors being limited to within $1$ meter in longitudinal/lateral shift and $1^\circ$ in angular error.

The contributions of this paper are two-fold:
\begin{itemize}
    \item a fine-grained cross-view localization method, SIBCL, that achieves accurate vehicle pose estimation with low spatial and angular errors.
    \item two branches of objectives design, RPRB encourages the predicted pose close to the ground truth, and PAB discriminates residuals across two views between the correct and incorrect pose.
\end{itemize}

\section{Related work}
\noindent\textbf{Visual Localization}.
Intensive research has been done in the field of visual localization for autonomous driving. 
SLAM methodologies \cite{mur2015orb,mur2017orb,engel2014lsd,davison2007monoslam,10.1109/ICRA.2017.7989741,jones2011visual,lynen2015get,mur2017visual,StephenWorst,schneider2018maplab,vora2019high} have traditionally been used for vehicle localization. They construct or update a geometrically accurate 3D map while simultaneously tracking the vehicle's location. The consecutive input sequence is essential since the geometrical accuracy relies on repeat observations of the same scene. As a result, SLAM methods are prone to error accumulation, which results in an estimation drift. In contrast, our method does not require a continuous series of images, and is able to estimate precise poses using a single query image paired with its corresponding 3D LiDAR points.
Other localization algorithms leave the mapping problem to an existing 3D scene model \cite{middelberg2014scalable,Sarlin_2019_CVPR, sarlin21pixloc,von2020lm} or directly utilize pre-collected 3D HD maps \cite{6942558}. 
Both the acquisition and maintenance of such a 3D model/map are laborious and expensive, especially for rural areas rarely visited by surveying vehicles.
 
\noindent\textbf{Cross-view Visual Localization}.
Satellite imagery is widely available, well-maintained, and easy to access. Recent works \cite{shi2020optimal,shi2019spatial,hu2018cvm,liu2019lending,toker2021coming,zhu2021vigor} resort to satellite images for localization.
All these methods estimate correspondences based on image-level features, making the localization inside a satellite image impossible. As a result, they do not perform well in fine-grained localization. 
Miller et al. \cite{9361130} proposed a cross-view SLAM method that uses semantically labeled LiDAR. Unlike our method, which focuses on localization-driven feature extraction, their approach relies heavily on semantic prior. Fervers et al. \cite{fervers2022continuous} proposed a cross-view localization method that ignores heading estimation and focuses on shift estimation. The most similar approach to ours is HighlyAccurate \cite{shi2020beyond}, a fine-grained cross-view localization method aimed at achieving 3-degree estimation.
It projects dense features from the satellite map onto the ground-view, under the assumption that all pixels in ground-view images are located on the ground plane. The features of above-ground pixels are projected to the incorrect position due to this wrong assumption, limiting its overall performance. 
In contrast, for more accurate pose estimation, our method builds geometric correspondences across reliable 3D LiDAR points. Furthermore, we adopt an attention map to reduce the influence of dynamic objects.

\section{Methodology} 
\begin{figure*}
    \centering
    \includegraphics[width=0.95\textwidth]{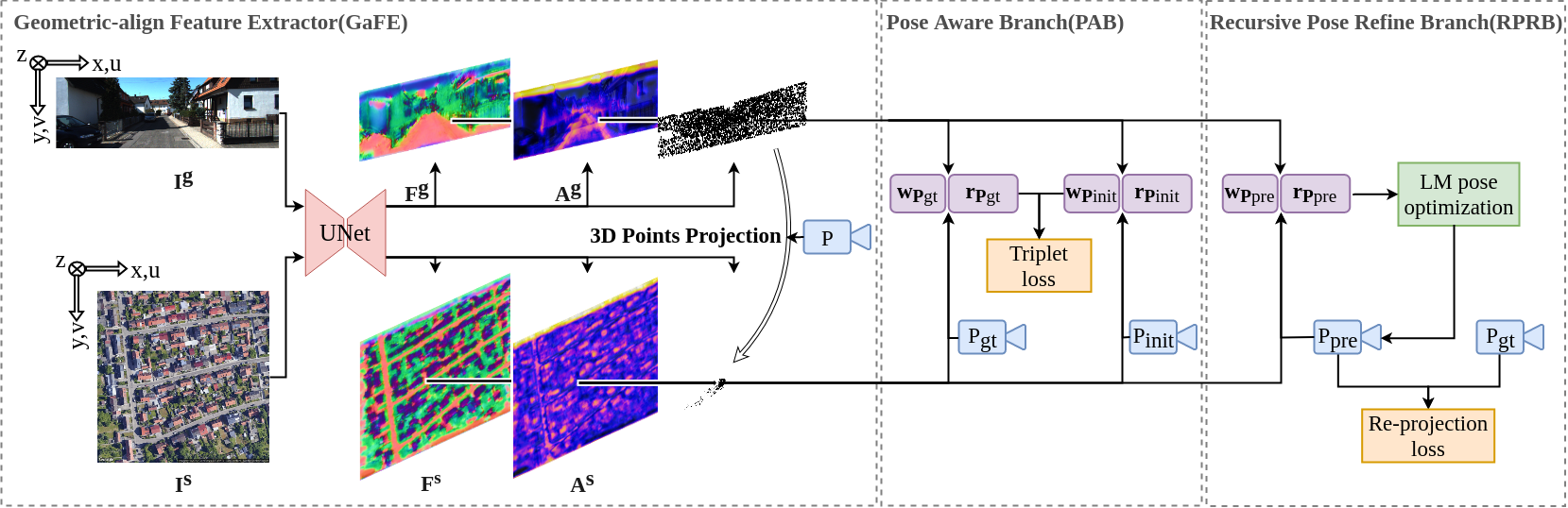}
    \caption{Overview of SIBCL. GaFE extracts feature and attention maps from the ground and satellite views. Further, we obtain sparse pixel-level features and weights across the 3D points. The pose-aware features extraction is supervised in PAB by a triplet loss. In RPRB, the deep features are used to iteratively optimize the camera pose using the LM algorithm.
    }
    \label{fig:architecture}
    \vspace{-0.2in}
\end{figure*}
\subsection{Overview}
Given a coarse pose
$\mathbf{P}_{init}$, our goal is to estimate the accurate 3-DoF pose $\mathbf{P}_{pre}$ of a ground-view camera
using a ground-view image $I^g$ and the 3D points in the ground-view (camera) domain $[x^g~ y^g~ z^g]^\top$. In our method, the 3D points are randomly sampled from valid LiDAR points.
The framework of the proposed SIBCL is illustrated in Fig.~\ref{fig:architecture}. The GaFE employs a Convolutional Neural Network (CNN) to extract feature maps $F^s$ and $F^g$ from the satellite and ground-view images, respectively. We adopt a U-Net structure of CNN that aims to obtain feature maps with original resolution that benefits accurate pose estimation. 
We also compute spatial attention maps $A^s$, $A^g$ to weight the feature maps, emphasizing pixels with potential correspondences between the two sets of images. 
Further, we project the 3D points onto the ground view images and satellite maps to obtain sparse pixel-level features and attention maps. Specifically, we obtain $F^g[p]$ and $A^g[p]$ from the ground-view images, and $F^s[p]_{\mathbf{P}}$ and $A^s[p]_{\mathbf{P}}$ from the satellite maps using the camera pose $\mathbf{P}$.
Residual $r_{\mathbf{P}}$ and point weights $w_{\mathbf{P}}$ are calculated from these sparse representations across the two views. In PAB, a triplet loss is employed to narrow the residual by the ground truth pose $r[p]_{\mathbf{P}_{gt}}$ and widen that by the initial pose
$r[p]_{\mathbf{P}_{init}}$. We only enable the PAB when the initial pose (incorrect pose) is significantly different from the ground truth pose. The RPRB is designed to iteratively optimize the predicted pose $\mathbf{P}_{pre}$ towards the ground truth pose $\mathbf{P}_{gt}$ using the LM algorithm. 

\subsection{Geometric-align Feature Extractor}
The GaFE extracts a hierarchy of ground-view and satellite feature maps at multiple resolutions, $F^{g/s} = \{F^{g/s}_{l} \in \mathbb{R}^{h_{l} \times w_{l} \times c_{l}} | l = 1, \dots, L\}$ where $l$ is the level of U-Net outputs, $h_{l}$, $w_{l}$, and $c_{l}$ represent the height, width, and channel number of feature maps in each level. Each pixel-level feature representation among these feature maps are $L_{2}$ normalised in order to improve the robustness to, e.g., variance in illumination conditions and viewpoints.
We further compute a spatial attention map $A^{g/s} = \{A^{g/s}_{l} \in \mathbb{R}^{h_{l} \times w_{l}} \}| l = 1, \dots, L\}$ by passing the un-normalized feature maps through a convolutional layer followed by a sigmoid activation function. This attention map is used to highlight pixels with potential cross-view correspondences. It assigns low score to temporal-inconsistent objects, like cars, and high scores to building edges and road marks that are identifiable from the satellite image. A visualization of the attention map is shown in Fig.~\ref{fig:confidence}.

\begin{figure}[!htb]
    \centering
    \begin{minipage}[b]{0.49\textwidth}
    \includegraphics[width=0.99\textwidth]{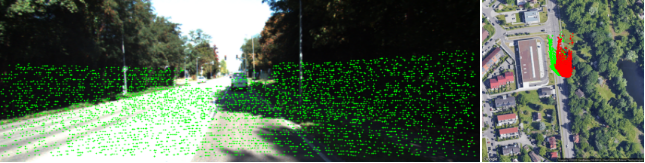}
    \end{minipage}
    \caption{(left) The visualizations of the projections of the 3D LiDAR points on the ground-view query image \textcolor{green}{(green)}. (right) The visualizations of the projections of the 3D LiDAR points on the satellite reference image using the initial pose \textcolor{red}{(red)} and the ground truth pose \textcolor{green}{(green)}, respectively.
    }
    \label{fig:points}
    \vspace{-0.1in}
\end{figure}

The coordinate systems of satellite and ground-view images are illustrated on the left of Fig.~\ref{fig:architecture}. The coordinates system of the satellite image is that the $x^s$-axis points to the east and the $y^s$-axis points to the south, and following the right-hand rule, the $z^s$-axis is vertically downward. The overhead-view satellite images are approximated as a parallel projection. The projection of 3D real-world objects onto a satellite image is formulated as:
\begin{equation}
    \begin{pmatrix}u^s\\v^s\end{pmatrix} = \begin{pmatrix}
    1/\gamma&0&c^s\\
    0&1/\gamma&c^s
    \end{pmatrix}\begin{pmatrix}x^s\\y^s\\1\end{pmatrix},
    \label{equ:s_intrinsic}
\end{equation}
where $(c^s, c^s)$ is the center of the satellite image and $\gamma$ is the meter-per-pixel ratio. 
\begin{equation}
    \gamma = \tilde{r}_{\text{earth}} \times \frac{\cos(\tilde{L} \times \frac{\pi}{180^{\circ}})}{2^{\tilde{z}} \times \tilde{s}},
    \label{equ:metre_per_pixel}
\end{equation}
where $\tilde{r}_{\text{earth}} = 156543.03392$ is the earth radius, $\tilde{L}$ is the latitude, $\tilde{z} = 18$ and $\tilde{s} = 2$ is the zoom factor and the scale of Google Maps \cite{google}, respectively. The projection of ground-view image is formulated as:
\begin{equation}
    [u^g~ v^g~]^\top \propto \mathbf{K}[x^g~ y^g~ z^g]^\top,
    \label{equ:g_intrinsic}
\end{equation}
where $\mathbf{K}$ is the intrinsic matrix of the ground camera. $[x^g~ y^g~ z^g]^\top$ is transformed from LiDAR domain to camera domain through the extrinsic matrix $\mathbf{M}_{lidar\rightarrow g}$.
The alignment between the ground and satellite 3D coordinate systems is calculated using:
\begin{equation}
    [x^s~ y^s~ z^s]^\top  = \mathbf{M}_{gps\rightarrow s}\mathbf{M}_{g\rightarrow gps}
    [x^g~ y^g~ z^g~ 1]^\top,
    \label{equ:extrinsic}
\end{equation}
where $\mathbf{M}_{g\rightarrow gps}$ is the extrinsic matrix from the camera pose to the GPS pose, given by calibration. $\mathbf{M}_{gps\rightarrow s}$ is the extrinsic matrix from the GPS pose to satellite coordinates, including three rotations: roll angle $\eta$, pitch angle $\vartheta$ and yaw angle $\theta$;  three translations: the lateral translation $\phi$, the longitudinal translation $\varphi$, and height translation fixed to GPS height. During pose optimization, roll and pitch are static, while yaw is optimized.
We look up sparse ground features and attention by projecting 3D points onto the correspondence views.
Fig.~\ref{fig:points} depicts the projection of 3D points onto the ground-view and satellite-view images. The projection points on satellite images depends on the pose of the query camera ($\phi$, $\varphi$ and $\theta$ to be estimated).

Residual $r[p]_{\mathbf{P}}$ is calculated by subtracting sparse features across two views.
\begin{equation}
r_{\mathbf{P}i} = F^s[\begin{pmatrix}u^g_i, v^g_i\end{pmatrix}^\top_{\mathbf{P}}] - F^g[\begin{pmatrix}u^s_i, v^s_i\end{pmatrix}^\top] \in \mathbb{R}^c,
\label{equ:risidual}
\end{equation}
where $[\cdot]$ is a lookup with sub-pixel interpolation. 
$c$ represents the feature dimension. 
The point weights $w_{\mathbf{P}}$ are the product of pixel-level attention across two views, for each 3D point i, we obtain:
\begin{equation}
w_{\mathbf{P}i} = A^s[(u^s_i,v^s_i)^\top_{\mathbf{P}}]\cdot A^g[(u^g_i,v^g_i)^\top] \in [0,1],  
\label{equ:D_pose}
\end{equation}
where $\cdot$ is element-wise product.

\subsection{Recursive Pose Refine Branch}
Having residuals and point weights, we use LM algorithm \cite{levenberg1944method,marquardt1963algorithm} to calculate the optimal solution of the vehicle pose $\mathbf{P}_{pre}$. The LM algorithm is solved by Cholesky decomposition in our approach.
We follow \cite{sarlin21pixloc} to optimize pose by successively utilizing features on each level, beginning with the coarsest level and initializing each level with the previous level's outcome. The Jacobian is defined as:
\begin{equation}
\mathbf{J}_i^{\delta} = \frac{\partial r_{\mathbf{P}i}}{\partial\delta}=\frac{\partial F^s}{\partial(u^s_i,v^s_i)_{\mathbf{P}}}\frac{\partial(u^s_i,v^s_i)_{\mathbf{P}}}{\partial(x^s_i,y^s_i)_{\mathbf{P}}}\frac{\partial(x^s_i,y^s_i)_{\mathbf{P}}}{\partial\delta},
\label{equ:Jacobian}
\end{equation}
where $\delta$ represents an update of each element in the 3-DoF pose.
The weight matrix $\mathbf{W}$ is constructed by stacking all points' weights to matrix diagonal:
\begin{equation}
\mathbf{W}=\text{Diag}(w_{\mathbf{P}i}\rho'),
\label{equ:Wight}
\end{equation}
where $\rho'$ is a derivative of the robust cost function $\rho$ \cite{hampel2011robust}. 
The update is calculated by damping the Hessian matrices $\mathbf{H}=\mathbf{J}^\top \mathbf{W}\mathbf{J}$ and solving the linear system as \eqref{equ:delta}.
\begin{equation}
\delta = -(\mathbf{H}+\lambda\text{ diag} (\mathbf{H}))^{-1}\mathbf{J}^\top \mathbf{W}\Upsilon,
\label{equ:delta}
\end{equation}
where 
$\Upsilon \in \mathbb{R}^{n\times c}$ is a matrix stacked of the residual $r_{\mathbf{P}i}$ described in \eqref{equ:risidual}, $n$ is the number of 3D LiDAR points. $\lambda$ is the damping factors \cite{sarlin21pixloc}.

\begin{figure}
    \centering
    \begin{minipage}[b]{0.45\textwidth}
        \includegraphics[width=0.99\textwidth]{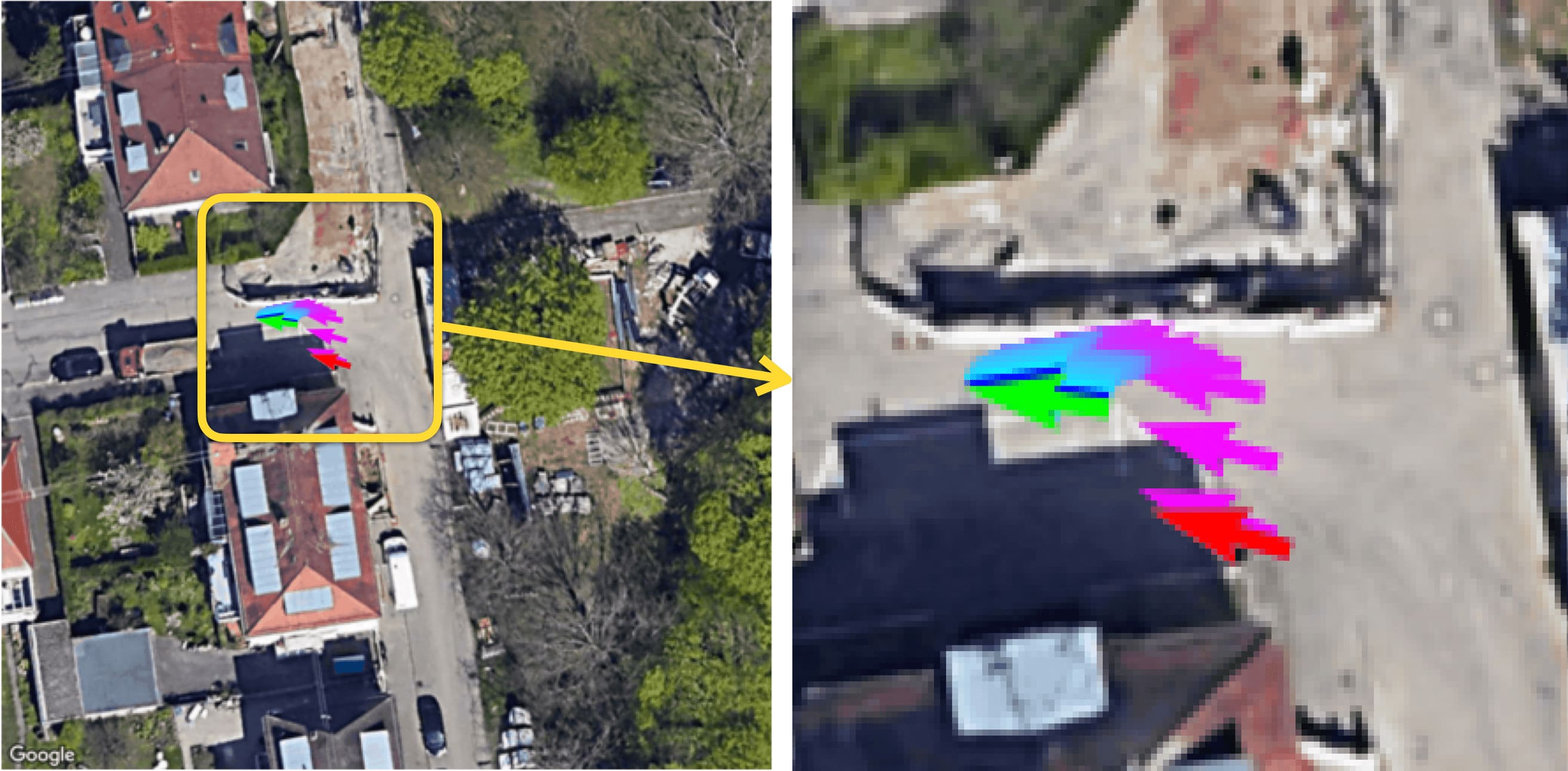}
    \end{minipage}
    \caption{The processing of pose optimization. RPRB optimizes the pose from an initial pose \textcolor{red}{(red)} towards the ground truth pose \textcolor{green}{(green)}. The final estimation pose \textcolor{blue}{(blue)}. The color of the estimated pose during the process changes from red to blue. 
    }
    \label{fig:pose_optimization}
    \vspace{-0.2in}
\end{figure}

The number of LM iterations is predetermined as $20$ throughout the training process.
During the test, the LM solver will stop when the update of all 3-DoF is less than $0.01$. We adopt a typical pose re-projection loss on the optimized pose. The re-projection loss is formulated as:
\begin{equation}
L_{RPRB}(\mathbf{P}_{pre}) =  \sum_i\|[u^s_{i}~ v^s_{i}]_{\mathbf{P}_{pre}}-[u^s_{i}~ v^s_{i}]_{\mathbf{p}_{gt}} \|_2^2,
\label{equ:reprojection_loss}
\end{equation}
where $()_{\mathbf{p}_{pre}}$ is 2D projection coordinates by the estimated pose, and $()_{\mathbf{p}_{gt}}$ is that by the ground truth pose. An example of the pose optimization process is shown in Fig.~\ref{fig:pose_optimization}. 

\subsection{Pose Aware Branch}
The PAB is designed to distinguish the correct pose from the erroneous ones in the feature space. This is achieved with a soft margin triplet loss \cite{qian2019softtriple}, defined as: 
\begin{equation}
L_{PAB} = log(1+e^{\alpha(1-\frac{\text{Dis}(\mathbf{P}_{init}))}{\text{Dis}(\mathbf{P}_{gt})}}),
\label{equ:triplet}
\end{equation}
$\alpha$ is a hyper-parameter that is empirically set to $10$. $\text{Dis}(\cdot)$ is a weighted distance that describes the distance of sparse features between ground and satellite views, formulated as:
\begin{equation}
\text{Dis}(\mathbf{P}) =\sum_{i}w_{\mathbf{P}i}\rho(\|r_{\mathbf{P}i}\|_2^2),  
\label{equ:D_pose}
\end{equation}
where $i$ is the index of 3D points, $\rho$ is a robust cost function \cite{hampel2011robust}. $r_{\mathbf{P}i}$ is the residual of 3D point i. $w_{\mathbf{P}i}$ is the point weight. Here, we use the initial pose as an erroneous pose. The initial pose can not be treated as a wrong pose if the initial pose is close to the ground truth. So we design a hyper-parameter weight $\beta$ for the triplet loss. We adopt PAB only if the re-projection error between the initial pose and ground truth pose $L_{RPRB}(\mathbf{P}_{init})$ is larger than a threshold that is empirically set to $10$. We also set a top-bound which is empirically set to $50$, to balance the impact of PAB.
\begin{equation}
L = L_{RPRB}(\mathbf{P}_{pre}) + \beta \times L_{PAB}, 
\end{equation}
where 
\begin{equation}
\beta = \begin{cases} 0& L_{RPRB}(\mathbf{P}_{init})<10\\ L_{RPRB}(\mathbf{P}_{init})& 10 \leq L_{RPRB}(\mathbf{P}_{init})\leq 50 \\ 50& L_{RPRB}(\mathbf{P}_{init})>50 \end{cases}.
\label{equ:loss}
\end{equation}
 
\section{Dataset}
We evaluate the proposed method in two standard autonomous driving datasets: KITTI \cite{geiger2013vision} and Ford Multi-AV Seasonal Dataset \cite{Agarwal_2020}. We construct KITTI-CVL and FordAV-CVL datasets by collecting the spatial-consistent satellite counterparts from Google Map \cite{google} according to the provided GPS tags. More specifically, we find the large region covering the vehicle trajectory and uniformly partition the region into overlapping satellite image patches. Each satellite image patch has a resolution of $1,280\times 1,280$ pixels, amounting to about $0.2$m per pixel of KITTI-CVL and $0.22$m per pixel of FordAV-CVL datasets. We discovered that the satellite images obtained from Google Maps sometimes shift slightly. In FordAV-CVL datasets, the satellite view sometimes does not match the ground view using the ground truth pose. After checking the six Ford Multi-AV Seasonal trajectories, we chose the `log4' trajectory for method evaluation, as `log4' has the best satellite view alignment. This misalignment is less severe in the KITTI-CVL dataset.
However, KITTI-CVL suffers from temporal misalignment. The camera images were taken over ten years ago, and some images are unmatched by those currently obtained satellite images due to the environment change.
We used all trajectories but removed some unmatched images. In future practices, the misalignment of satellite maps can be avoided by using more accurate commercial satellite maps. We use images obtained by the front left camera from both datasets as our query inputs.

\noindent\textbf{Training, Validation and Test Sets}.
The KITTI \cite{geiger2013vision} data contains various trajectories captured at different times, with little overlap in the trajectories captured. Our validation and training data are from the same trajectory because the validation sets are used to select the best-performing model during training. In contrast, the test sets are from different trajectories for the generalization ability evaluation. For Ford Multi-AV Seasonal \cite{Agarwal_2020}, each trajectory has been experienced three times with different weather, lighting, and traffic conditions. We aim to train our approach on one trajectory and test on the same trajectory but at a different time under different conditions. 

\section{Experiments}
\noindent\textbf{Metrics}.
Our goal is to estimate the 3-DoF pose.
we report the median errors in 
lateral and longitudinal translation (m), yaw rotation ($^\circ$) errors, and also the localization recall under thresholds ($0.25$m, $0.5$m, $1$m, $2$m, $1^\circ$, $2^\circ$, $4^\circ$). Since the satellite image is about $0.2$m per pixel, We choose $0.25$m shift error as the minor error range.

\noindent\textbf{Implementation Details}.
Because RTK signals have already corrected the provided GPS tags in both KITTI \cite{geiger2013vision} and Ford Multi-AV Seasonal \cite{Agarwal_2020} datasets, we use them as ground truth pose. 
Unless specifically stated, the initial pose is randomly sampled under $30^\circ$ yaw angle errors and $10$m lateral and longitudinal shifts based on the provided GPS throughout the experiments. 
The LiDAR data in KITTI \cite{geiger2013vision} have been synchronized with camera images, we randomly sample 5000 \footnote{We chose this sample size for training based on our experience. We have since discovered that increasing the number of points beyond this threshold does not result in significant improvements.} valid raw LiDAR points for each ground-view image processing. Raw LiDAR data is not synchronized with camera images in Ford Multi-AV Seasonal Dataset \cite{Agarwal_2020}. So, we randomly sample 5000 3D points that overlap with the camera image from the 3D map instead.
VGG19 pre-trained on ImageNet \cite{deng2009imagenet} is adopted as an encoder to construct our U-Net. 
We adopt batch size $b=3$ and Adam optimizer \cite{kingma2014adam} with a learning rate of $10^{-5}$. 

\noindent\textbf{Inference Speed}.
The processing time of the GaFE is around $150$ms. The optimization process is executed for $20$ iterations at each level, and it takes about $200$ms in total.

\begin{table*}[!htb]
\renewcommand{\arraystretch}{1.0}
\caption{Cross-view Methods Comparison} 
\label{Tab:SOTA_Comparison}
\centering
\setlength\tabcolsep{1pt}
\begin{tabular*}{\textwidth}{c c||@{\extracolsep{\fill}}c c c c c|c c c c c|c c c c}
\hline
&  & \multicolumn{5}{c|}{\bfseries Lateral} & \multicolumn{5}{c|}{\bfseries Longitudinal} & \multicolumn{4}{c}{\bfseries Yaw} \\
& & median$\downarrow$ & 0.25m$\uparrow$ & 0.5m$\uparrow$ & 1m$\uparrow$ & 2m$\uparrow$ & median$\downarrow$ & 0.25m$\uparrow$ & 0.5m$\uparrow$ & 1m$\uparrow$ & 2m$\uparrow$ & median$\downarrow$ & $1^\circ\uparrow$ & $2^\circ\uparrow$ & $4^\circ\uparrow$ \\
\hline\hline
\multirow{3}{*}{\makecell[c]{KITTI-CVL\\(evaluation)}} 
&\bfseries DSM\cite{shi2020looking} & 2.62 & 10.09 & 15.41 & 25.89 & 42.36 & 3.66 & 9.96 & 13.81 & 24.04 & 35.59 & 11.87 & 3.92 & 9.18 & 18.32\\
&\bfseries HighlyAccurate\cite{shi2020beyond} & 0.63 & 21.73 & 41.24 & 68.38 & 87.51 & 1.97 & 7.55 & 14.34 & 27.35 & 50.62 & 1.40 & 38.21 & 62.21 & 81.29\\
&\bfseries Ours & \textbf{0.22} & \textbf{54.38} & \textbf{82.54} & \textbf{95.35} & \textbf{97.83} & \textbf{0.23} & \textbf{52.03} & \textbf{77.95} & \textbf{92.02} & \textbf{96.33} & \textbf{0.46} & \textbf{81.83} & \textbf{95.01} & \textbf{98.65}\\
\hline
\multirow{3}{*}{\makecell[c]{KITTI-CVL\\(test)}} 
&\bfseries DSM\cite{shi2020looking} & 3.48 & 8.51 & 13.97 & 23.44 & 40.61 & 4.97 &7.67 & 11.02 & 20.03 & 30.48 & 12.73 & 3.13 & 8.29 & 17.44\\
&\bfseries HighlyAccurate\cite{shi2020beyond} & 0.83 & 16.51 & 32.05 & 57.65 & 83.11 & 2.01 & 7.14 & 14.11 & 27.41 & 49.94 & 1.82 & 29.83 & 53.41 & 76.51\\
&\bfseries Ours & \textbf{0.54} & \textbf{25.59} & \textbf{46.26} & \textbf{72.63} & \textbf{89.78} & \textbf{0.64} & \textbf{21.91} & \textbf{41.22} & \textbf{64.47} & \textbf{80.37} & \textbf{0.85} & \textbf{56.05} & \textbf{79.70} & \textbf{90.89} \\
\hline
\multirow{3}{*}{\makecell[c]{FordAV-CVL\\(test)}} 
&\bfseries DSM\cite{shi2020looking} & 3.86 & 8.23 & 12.47 & 18.93 & 29.24 & 5.01 & 6.20 & 10.25 & 16.39 & 27.28 & 12.03 & 3.76 & 8.92 & 17.36\\
&\bfseries HighlyAccurate\cite{shi2020beyond} & 0.84 & 16.56 & 31.31 & 57.64 & 85.45 & 1.82 & 7.11 & 13.87 & 28.53 & 53.64 & 1.83 & 30.74 &  53.08 & 78.40\\
&\bfseries Ours & \textbf{0.55} & \textbf{24.83} & \textbf{45.90} & \textbf{74.06} & \textbf{89.14} & \textbf{0.78} & \textbf{18.72} & \textbf{34.11} & \textbf{58.26} & \textbf{75.44} & \textbf{0.57} & \textbf{66.76} & \textbf{81.78} & \textbf{90.50} \\
\hline
\multirow{2}{*}{\makecell[c]{Cross-Datasets\\KITTI$\rightarrow$FordAV}} 
&\bfseries HighlyAccurate\cite{shi2020beyond} & 3.17 & 4.02 & 8.45 & 16.76 & 33.59 & 3.11 & 4.56 & 8.57 & 17.09 & 32.86 & 6.59 & 8.15 &  16.28 & 32.29\\
&\bfseries Ours & \textbf{1.27} & \textbf{10.93} & \textbf{21.04} & \textbf{40.84} & \textbf{65.58} & \textbf{1.63} & \textbf{8.59} & \textbf{16.97} & \textbf{32.28} & \textbf{57.75} & \textbf{1.67} & \textbf{32.65} & \textbf{56.20} & \textbf{73.45} \\
\hline
\end{tabular*}
\vspace{-0.2in}
\end{table*}

\noindent\textbf{Comparison with Image Retrieval Method DSM \cite{shi2020looking}}. 
In order to adapt DSM \cite{shi2020looking} method to our task, we retrieve the ground-view image in a subset of satellite images. The subset of satellite images is constructed by cropping one big correspondence satellite image with $0.25$m center shift on both $u^s$-axis and $v^s$-axis. For example, with 5m$\times$5m shift range, we need to crop $20 \times 20$ $(5/0.25) = 400$ satellite images as reference images for a ground-view query image. Then, we feed the one ground-view image and $400$ cropped satellite images into the DSM network to retrieve the most similar satellite image. We calculate the lateral and longitudinal distance between this retrieved satellite image center and the ground truth pose. The evaluation results are reported in Tab.~\ref{Tab:SOTA_Comparison}. It is clear that our method outperforms DSM \cite{shi2020looking} with a large margin. When reference images are close to each other, it is hard for image retrieval methods to distinguish image-level differences.

\noindent\textbf{Comparison with Fine-grained Localization Method HighlyAccurate \cite{shi2020beyond}}.
We train HighlyAccurate \cite{shi2020beyond} model in the same setting. The evaluation results are reported in Tab.~\ref{Tab:SOTA_Comparison}. Our method significantly outperforms HighlyAccurate \cite{shi2020beyond} in both KITTI-CVL and FordAV-CVL datasets, especially on longitudinal estimation, even though our memory usage ($4736$MB) is less than HighlyAccurate ($6445$MB). HighlyAccurate \cite{shi2020beyond} does not ignore dynamic objects. It
constructs geometric correspondence on the assumption that all pixels of camera images lie on the ground plane. This unreasonable assumption limits its performance.
Besides, our approach provides correct geometric correspondence across 3D LiDAR points. The sparse feature alignment is reliable and efficient. To improve robustness and accuracy, we also adopt attention maps to reduce the impact of dynamic objects.

\noindent\textbf{Generality}.
Our method demonstrated good generalization capabilities for new scenes, as shown in Tab.~\ref{Tab:SOTA_Comparison}. 
\enquote{FordAV-CVL(test)} images are from the same trajectories but different drives of the training dataset, with different environmental conditions, e.g., different weather and viewpoints. \enquote{KITTI-CVL(test)} images are from different trajectories (unseen since). \enquote{Cross-Datasets} is more challenging. We train the model in \enquote{KITTI-CVL} and test it on \enquote{FordAV-CVL(test)}. In addition to different trajectories, the camera setting is also different. We compare our cross-datasets generality with HighlyAccurate \cite{shi2020beyond}, and the result illustrates our generalization to strongly differing scenes and settings due to the benefits from sparse feature alignment and attention map mechanisms.

\noindent\textbf{Attention Map Visualization}.
\begin{figure}[!htb]
    \centering
    \begin{minipage}[b]{0.40\textwidth}
        \includegraphics[width=0.99\textwidth]{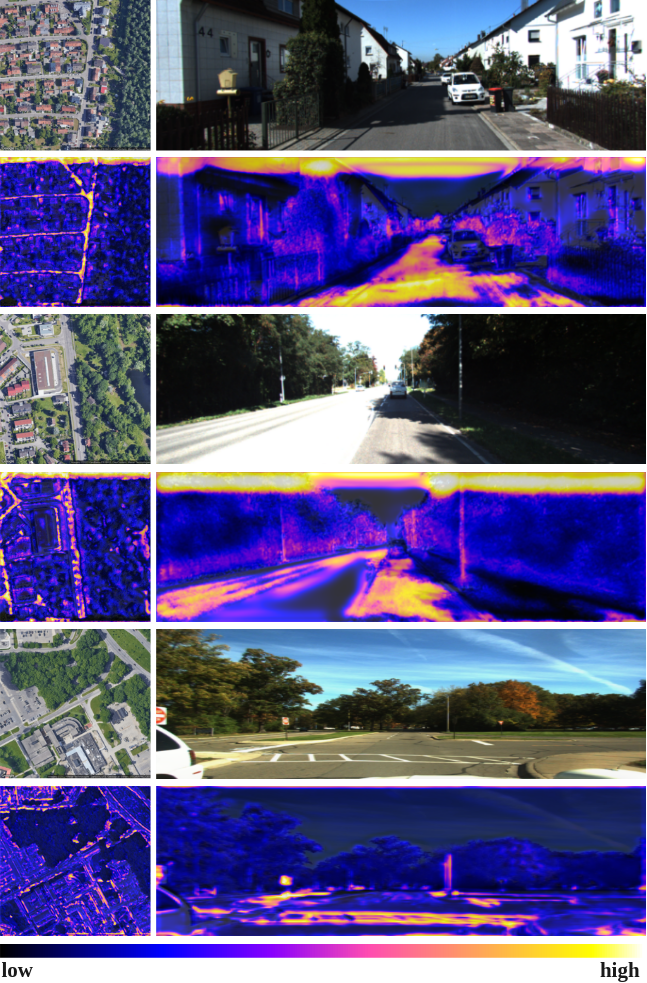}
    \end{minipage}
    \caption{Illustration of satellite, ground-view images, and their corresponding attention maps. $1^{st}$ and $2^{nd}$ are from the KITTI-CVL dataset. $3^{rd}$ is from the FordAV-CVL dataset. 
    }
    \label{fig:confidence}
    \vspace{-0.2in}
\end{figure}
We demonstrate attention maps in Fig.~\ref{fig:confidence}, which is helpful for us to discover which cues are useful or detrimental for localizing in ground-view and overhead-view images. Our network successfully extracts semantic features, e.g., road masks and edges, the boundaries of buildings and poles, etc., which contribute greatly to vehicle localization. Moreover, moving objects and repeated objects, e.g., vehicles and leaves of trees, with negative impact, are ignored. For the ground view image, it is noteworthy that the sky is assigned a high score because it is beyond the height range of the LiDAR points and is not sampled in the algorithm. As the sky's confidence score is not actively monitored during training, it may retain an initial high value. However, given that the sky does not contribute to pose optimization, its high confidence score will not have any impact on vehicle localization.
Additionally, our method employs a confidence mechanism that eliminates the requirement for simultaneous acquisition of satellite and ground-view imagery. This mechanism automatically reduces the impact of dynamic objects and prioritizes more stable objects that are more useful in vehicle pose estimation.

\begin{figure}[!htb]
    \centering
    \subfloat[KITTI-CVL(test)]{
        \begin{minipage}[t]{0.45\textwidth}
            \includegraphics[width=0.49\textwidth]{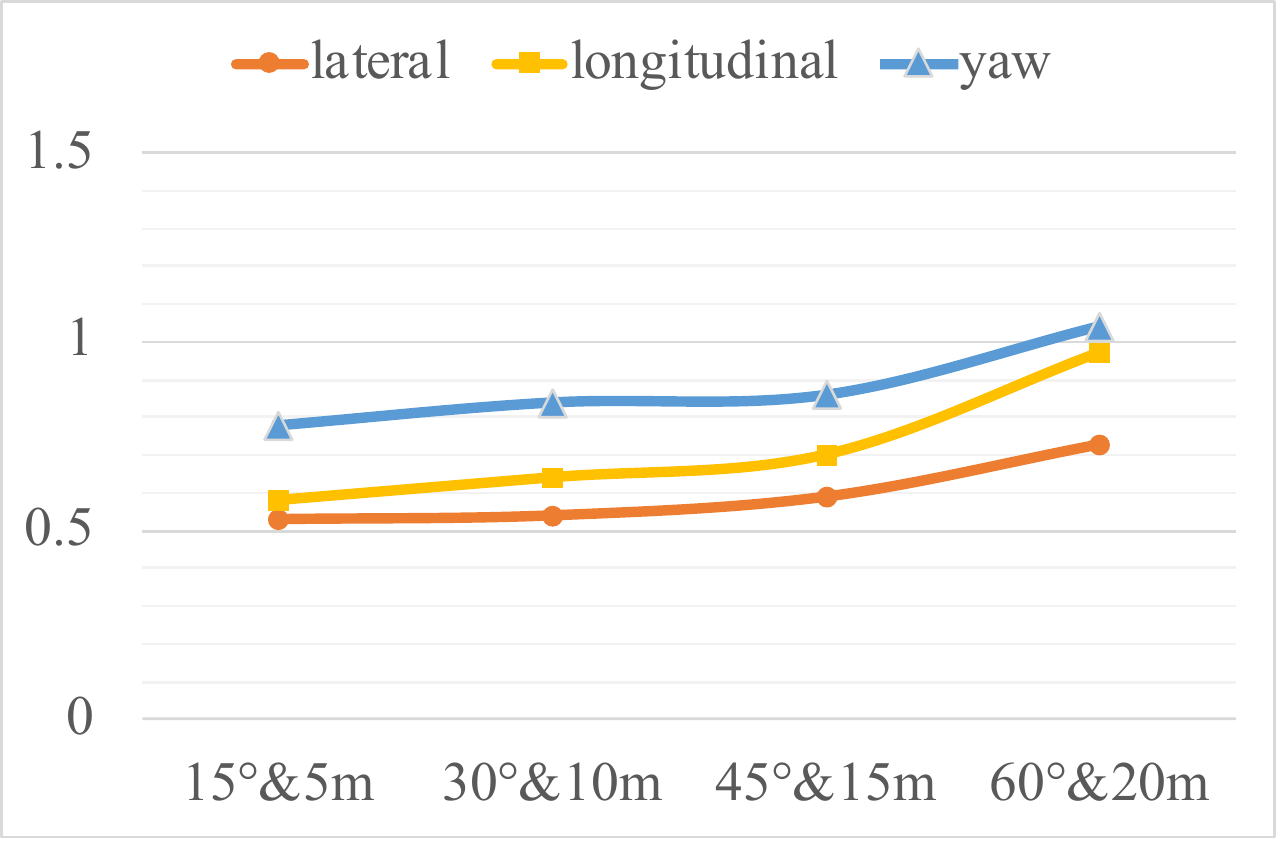}
            \includegraphics[width=0.49\textwidth]{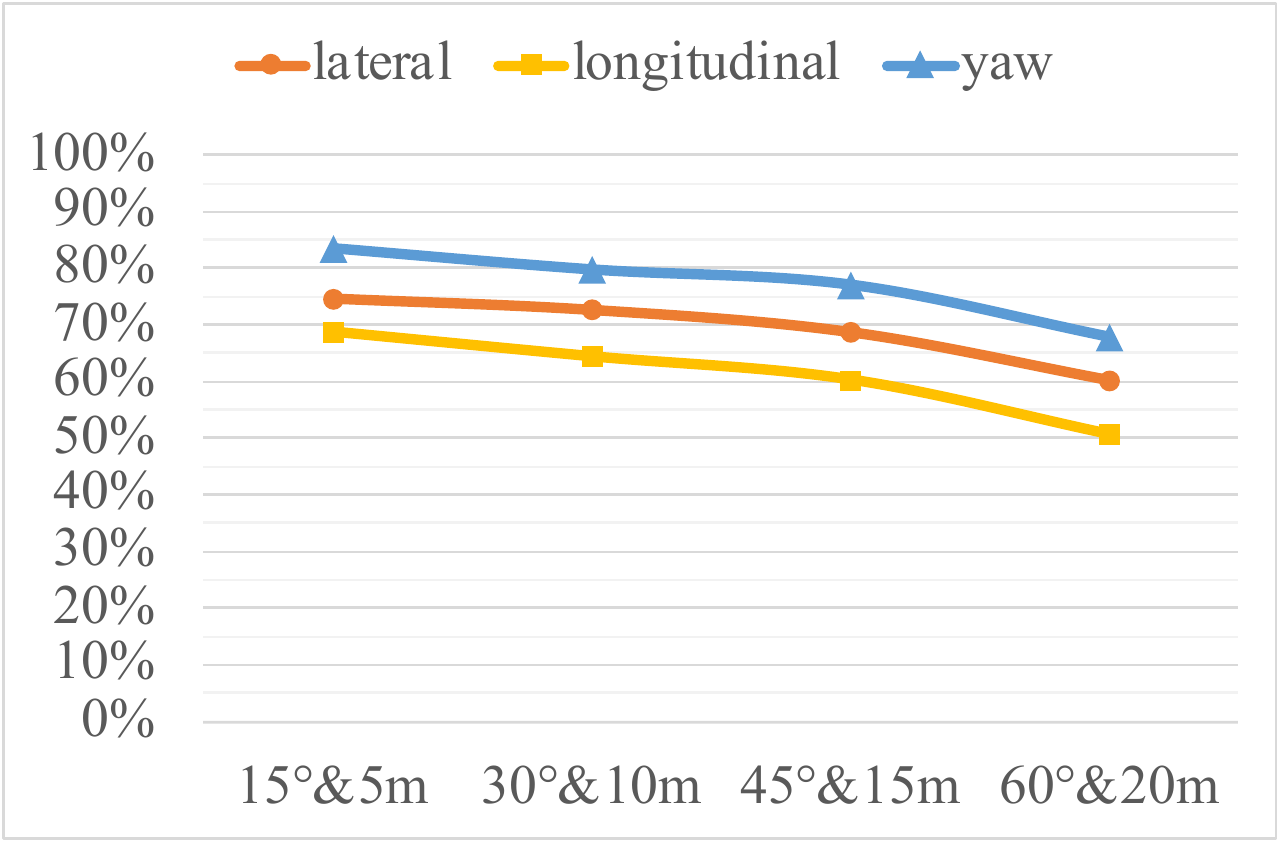}
        \end{minipage}
    }
    \\
    \subfloat[FordAV-CVL(test)]{
        \begin{minipage}[t]{0.45\textwidth}
        \includegraphics[width=0.49\textwidth]{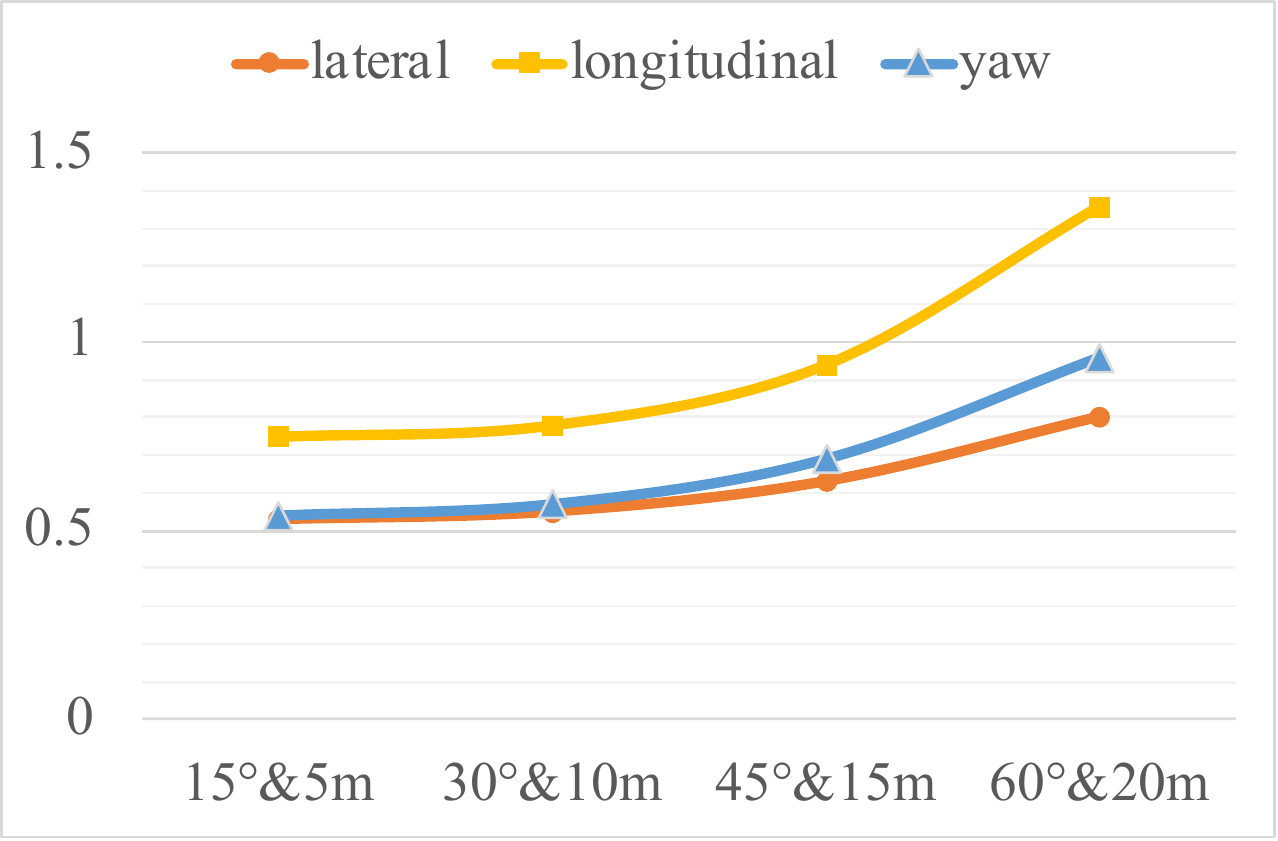}
        \includegraphics[width=0.49\textwidth]{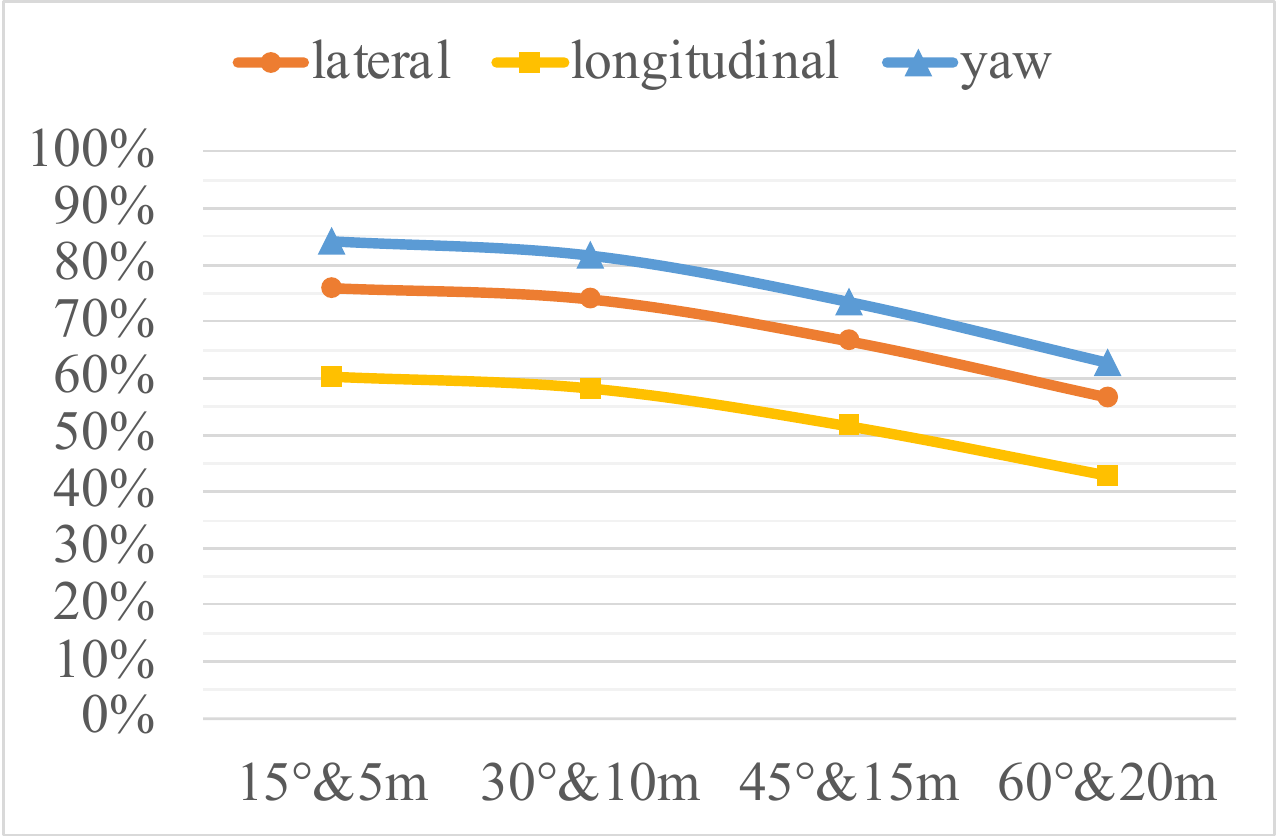}
        \end{minipage}
    }
    \caption{(left) Median lateral and longitudinal translation (m) and yaw rotation ($^\circ$) errors under different initial poses. (right) Recall under lateral and longitudinal translation error 1m and yaw rotation errors $2^\circ$.}
    \label{fig:median}
    \vspace{-0.2in}
\end{figure}

\noindent\textbf{Performance under Different Initial Poses}.
We tested our method using the initial pose with a more extensive and challenging bound. The results are shown in Fig.~\ref{fig:median}. It shows that our method not only performs well with the initial poses under the same level when training the network but is also robust when using the initial pose under a larger shift. To be more specific, our approach is robust and achieves a satisfactory accuracy
with initial raw pose under $60^\circ$ yaw angle errors and $20$m lateral and longitudinal shifts.

\section{Ablation Study}
\noindent\textbf{Shared Weights between the Two Views}.
One of the key challenges in cross-view visual localization is to extract features that are robust to the appearance gap between ground-view and overhead-view images.
Satellite and ground-view images have different resolutions, different viewpoints, and various camera intrinsic. That is why most cross-view methods \cite{hu2018cvm,shi2020optimal,shi2019spatial,shi2020looking,zhu2021vigor,toker2021coming, shi2020beyond} use siamese architecture without shared weights to do metric learning. In contrast, \cite{sarlin21pixloc} shares weights between two branches and adopts attention maps to mitigate domain shift. We evaluate both methods, and the result of the KITTI-CVL (test) is shown in Tab.~\ref{Tab:Ablation_study}. The latter (full) performs much better than the former (w/o SW). A potential reason is that the shared weights branches help the learned features be in the same domain. Thus, shared weights facilitate more accurate pose estimation.

\begin{table}[!htb]
\caption{Ablation Study}
\label{Tab:Ablation_study}
\centering
\renewcommand{\arraystretch}{0.95}
\begin{tabular}{c||@{\extracolsep{\fill}}cc|cc|cc}
\hline
& \multicolumn{2}{c|}{\bfseries Lateral} & \multicolumn{2}{c|}{\bfseries Longitudinal} & \multicolumn{2}{c}{\bfseries Yaw} \\
& median$\downarrow$ & 1m$\uparrow$ & median$\downarrow$ & 1m$\uparrow$ & median$\downarrow$ & $2^\circ\uparrow$ \\
\hline
\bfseries w/o SW & 0.84 & 56.25 & 1.19 & 44.32 & 1.51 & 58.88  \\
\bfseries w/o PAB & 0.59 & 69.64 & 0.68 & 62.89 & 0.86 & 78.74 \\
\bfseries w/o $\beta$ & 0.58 & 70.10 & 0.68 & 64.04 & 0.85 & 79.20 \\
\bfseries full & \textbf{0.54} & \textbf{72.63} & \textbf{0.64} & \textbf{64.47} & \textbf{0.85} & \textbf{79.70}\\
\hline
\end{tabular}
\vspace{-0.1in}
\end{table}

\noindent\textbf{Effectiveness of PAB}.
We designed two branches of objective functions. The RPRB encourages the predicted pose close to the ground truth. 
The PAB discriminates residuals 
between the correct and incorrect pose. Unlike RPRB which passes gradient from pose to feature extractor weights, the gradient backpropagates of PAB are directly from triplet loss to feature extractor weights. The PAB allows the network to learn the pose-aware features before RPRB can estimate a proper pose. 
We adopt a hyper-parameter PAB loss weight $\beta$ to balance the effect of PAB. 
The results of the ablation studies on $\beta$ and PAB of KITTI-CVL(test) are shown in Tab.\ref{Tab:Ablation_study}. They all contribute to the performance of SIBCL. 



\section{Conclusions}
Our paper presents a robust and novel  geometry-driven correspondence learning approach for 3-DoF camera pose estimation. 
SIBCL is the first cross-view approach capable of accomplishing localization with a median position error below $1$ meter and a median orientation error below $1^\circ$, without relying on a high-definition map.
It makes maximum use of satellite images for fine-grained localization. 
Furthermore, it generalizes well to new scenes and thus can be used as an interpretable prior and adapted to new scenes after brief fine-tuning.
In the future, we will consider using multi-cameras and low-cost sensors. We believe our work may lead to reliable, accurate, and low-cost vehicle localization systems.







\bibliographystyle{./bibliography/IEEEtran}
\nocite{*}
\bibliography{./bibliography/IEEEabrv}

\end{document}